\documentclass[10pt,letterpaper]{article}
\usepackage{color}

\usepackage{cogsci}
\usepackage{pslatex}
\usepackage{apacite}
\usepackage{marvosym}
\usepackage{amsmath,amssymb,amsfonts}
\usepackage{graphicx}
\usepackage{epsf}

\newcommand{\noiseBOX}{\includegraphics[width=.035\textwidth]{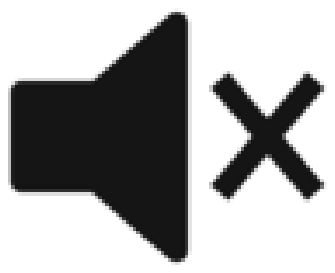} }
\newcommand{\noiseS}{\includegraphics[width=.021\textwidth]{mute.eps}}

\newcommand{\as}{\mathfrak{s}}
\newcommand{\asT}{{$\mathfrak{s}$ }}
\newcommand{\asTe}{$\mathfrak{s}$}
\title{Modeling the Ellsberg Paradox by Argument Strength}
 
\author{{\large \bf Niki Pfeifer (niki.pfeifer@lmu.de)} \\
Munich Center for Mathematical Philosophy, LMU Munich, Germany
  \AND {\large \bf Hanna Pankka (hanna.pankka@helsinki.fi)} \\
   Department of Philosophy, History, Culture and Art Studies, University of Helsinki, Finland}

\begin{document}

\maketitle

\begin{abstract}
We present a formal measure of argument strength, which combines  the ideas that  conclusions of strong arguments are (i) highly probable  and (ii) their uncertainty is  relatively precise. Likewise, arguments are weak when their conclusion probability is low or when it is highly imprecise. We show how the proposed measure provides a new model of the Ellsberg paradox. Moreover, we further substantiate the  psychological plausibility of our approach  by   an experiment ($N=60$). The data   show that the proposed measure predicts human inferences in the original Ellsberg task and in corresponding argument strength tasks. Finally, we report qualitative data taken from structured interviews on folk psychological conceptions on what argument strength means.

\end{abstract}

\section{Introduction}

\subsection{Measuring Argument Strength}
\label{SEC:IntroMeasure}
Probabilistic models of argumentation became popular in cognitive science and its subfields including psychology, philosophy, and computer science in recent years \cite<see, e.g.,>{hahn06a,haenni09,zenker13}. Like logic-based nonmonotonic approaches for defeasible argumentation \cite<see, e.g.>{prakken02}, probabilistic approaches allow for dealing with exceptions and retracting conclusions in the light of new evidence. However, in contrast to qualitative logical approaches, probability allows for managing \emph{degrees of belief} in the sentences involved in common sense argumentation. Moreover, degrees of belief can be used to   model  the strength of arguments \cite{hahn06a,oaksford07c,pfeifer06d,pfeifer07,pfeifer13a}. 

The concept ``argument'' is ambiguous. In logic, it denotes a triple consisting of a (possibly empty) premise set, a conclusion indicator, and a conclusion set. Consider, for example, the following argument, which is an  instance of modus ponens:
\begin{tabular}{ll}
(P1)&If I take the train at five ($T$), I'll be home at six ($H$).\\
(P2)&I take the train at five ($T$).\\
(C) &Therefore, I'll be home at six ($H$).\\
\end{tabular} 
Here, (P1) and (P2) are the premises, ``Therefore'' the conclusion indicator and the sentence ``I'll be home at six'' is the conclusion.  In argumentative contexts, ``argument'' may also denote a premise which speaks for or against a conclusion. For example ``The train conductors are on strike'', can serve as an argument for concluding that it is better to take the bus. In what follows, however, we will focus on arguments in the logical sense only.

How can we measure the strength of an argument? There are at least two formal approaches to study (probabilistic) argument strength.  In the first approach argument strength is    based on \emph{uncertain consequence} relations, i.e., by presupposing that the conclusion follows to some degree from the premises. Usually, this is modeled by a conditional probability of  ``the conclusion \emph{given} (some combination of) the premises'' of the argument \cite<see, e.g.>{hahn06a,oaksford07c}. As pointed out by \citeA{osherson90}, measures of confirmation can serve as models for argument strength \cite<for an overview of measures of confirmation see>{crupi07}. Measures of confirmation and previous attempts to model argument strength by uncertain consequence relations are problematic when arguments involve conditionals, like the modus ponens above  (see premise (P1)): it is far from clear to give a precise meaning of conditionalizing on a combination of premises, when the premise set contains conditional events. There is ample formal and experimental evidence that uncertain conditionals are best modeled by conditional probabilities \cite<see, e.g.,>{evans04,oaksford07,OverCruz17,pfeifer13,pfeifer13b}. Therefore, conditionals should be modeled by conditional probabilities. However, this  requirement would imply to measure the uncertainty of a \emph{conclusion \underline{given} (some combination of) the premises}. Unfortunately, satisfactory  semantics of expressions like $\overbrace{C}^{\text{conclusion}} | \overbrace{(A \text{ and } (C|A))}^{\text {premises}}$ do not exist yet. Such semantics would, however, be necessary to capture the underlying logical structure of the modus ponens \cite<for an approach where conditionals are interpreted as conditional random quantities which allows for dealing with nested conditionals, see>{2016:SMPS1,GOPSsubm}. Modus ponens is just a relatively simple example here: there are, of course, many other argument forms involving conditionals.  The inability to deal with conditionals seems to us to be one of the main reasons, why currently no formally satisfactory measure of argument exists within the first approach: measures based on uncertain consequence relations do not seem to be able to deal with the logical form of the argument. 

In this paper, we advocate the second approach to argument strength. It satisfies the requirement of doing justice to the logical form of arguments involving conditionals \cite{pfeifer07,pfeifer13a}. Specifically, we define argument strength based on the following ideas: (i) keep the consequence relation deductive, (ii)  assign probabilities to the premises, and then (iii) define the measure of argument strength based on the  propagated coherent lower and upper probability bounds on the conclusion \cite{pfeifer07,pfeifer13a}. 
Probability propagation from the premises to the conclusion is governed by   \emph{coherence based probability logic} \cite<see, e.g.>{coletti02,pfeifer09b,gilio16}. The  \emph{coherence approach} to probability was originated by Bruno de Finetti \cite{definetti74}. It  conceives probabilities  as subjective degrees of belief. Conditional probabilities ($p(C|A)$) are primitive. This allows for zero probabilities of the conditioning event ($A$). Note that in standard approaches to probability,  $p(C|A)$ is undefined if $p(A)=0$, which is problematic in many argument forms \cite<see, e.g.>{pfeifer13,gilio16}.
Moreover, coherence allows for managing \emph{imprecise probabilities} (set-valued probabilities involving lower and upper probability bounds), which is relevant for formalising arguments under incomplete probabilistic knowledge. The above mentioned modus ponens, for example, is formalised as follows:\\
\begin{tabular}{lp{7cm}}
(P1') & $p(H|T)=x$\\
(P2') &$p(T)=y$\\
(C') & Therefore, $z' \leq p(H)\leq z''$, where $z'=xy$ and $z''=xy+1-y$  are the best possible coherent probability  bounds on the conclusion.\\ 
\end{tabular}\\
Following \citeA{pfeifer13a}, we define the measure of argument strength \asT on an argument $\mathcal{A}$ as follows:
\begin{quote}
Let $z'$ and $z''$ denote the coherent lower and upper probability bounds, respectively,  on the conclusion of argument $\mathcal{A}$. Then,
\begin{equation}\label{EQN:ArgStr}
\as(\mathcal{A})=_{\text{def.}}\overbrace{(1-(z''-z'))}^{\text{precision}} \times \overbrace{\frac{z'+z''}{2}}^{\text{location}}\, .
\end{equation}
\end{quote}
 Intuitively, measure \asT  combines the \emph{precision} and the \emph{location} of the coherent conclusion probability interval. Specifically, strong arguments are arguments with \emph{low imprecision} of the conclusion probability (measured by the one-complement of the distance between the upper and the lower probability bounds, $1-(z''-z')$) and with conclusion probabilities \emph{close to one} (measured by the mean of the lower and upper probability bound, $(z'+z'')/2$). Of course, precision and location could be modeled differently (e.g., by using the geometric or the harmonic mean instead of the arithmetic mean). Moreover, in contexts where the location is more important than the precision of the conclusion probability interval (or \emph{vice versa}), adding suitable weights to formula~(\ref{EQN:ArgStr}) can  adjust the measure for such cases. However, for the purpose of our paper it is sufficient to keep the measure as simple as possible.

Measure \asT has a number of plausible consequences:
it  ranges always from zero to one (i.e., $0 \leq \as\leq 1$, since $z'$ and $z''$ are probability values, which are also in the unit interval, $[0,1]$). The extreme ``0'' denotes weak arguments and ``1'' denotes strong arguments. Arguments with conclusion probability  1, are strong arguments, since $\as = 1$ if $z'=z''=1$. Arguments with conclusion probability 0 (i.e., $z'=z''=0$) are weak arguments, since  $\as = 0$. Likewise, probabilistically non-informative arguments (i.e., $z'=0$ and $z''=1$) are weak arguments, since   $\as = 0$. 

Interestingly, measure \asT also provides a new solution to the Ellsberg paradox \cite{ellsberg61},\footnote{We thank Kevin T. Kelly for pointing us to the Ellsberg paradox.} which we describe in the next section.
\subsection{Modeling the Ellsberg Paradox by Measure \asT}
\label{SEC:IntroEllsberg}

 Ellsberg described the following situation \cite{ellsberg61}: 
\begin{quote}
  An urn   contains 90 balls, of which 30 are red ($R$) and 60 are black or yellow ($B \vee Y$, where ``$\vee$'' denotes \emph{disjunction} (``or'') as defined in classical logic). The ratio of the black and yellow balls is unknown---there might be anything between 0 to 60 black (or yellow) balls. One ball is drawn from the urn and you are asked to choose a bet between two bets. If you take \textbf{Bet 1}, you will win \$100, if the ball drawn from the urn is red. If you take \textbf{Bet 2}, you will win \$100, if the ball drawn from the urn is black. 
\end{quote}
Ellsberg predicted that  most people choose {\bf Bet 1} when asked to decide which of the two bets they prefer. Then, considering again the same urn, Ellsberg predicted that people will choose {\bf Bet 4}, when they are asked to decide between the following two alternative bets: 
\begin{quote}
If you take {\textbf{Bet 3}}, you will win \$100, if the ball drawn from the urn is red or yellow. If you take {\textbf{Bet 4}}, you will win \$100, if the ball drawn from the urn is black or yellow.
\end{quote}
Ellsberg's predictions create a well-known paradox as they violate the independence axiom of rational choice  \cite<see, e.g.,>{sep-rationality-normative-utility}. Moreover, Ellsberg's predictions were experimentally confirmed in many studies \cite<see, e.g.,>{becker64,slovic74,maccrimmon79}.

We propose to frame the Ellsberg paradox in terms of probability logical arguments. Specifically, the \emph{premises} represent the probabilistic information given in the description of the urn, and the \emph{conclusions} represent the respective bets involved in the Ellsberg paradox.  Thus, we obtain four arguments. Each argument speaks for choosing the corresponding bet.  The associated argument to \textbf{Bet 2}, for example, is argument $\mathcal{A}_2$:
\begin{quote}
$p(R)=.33$\\  $p(B \vee Y)=.67$\\
Therefore, $0\leq p(B)\leq.67$ is coherent.
\end{quote} 
The strength of this argument is denoted by  $\as(\mathcal{A}_2)$ and by applying equation~(\ref{EQN:ArgStr}) equal to .11 (i.e., $\as(\mathcal{A}_2)=.11$).
Table~\ref{TAB:ArgEllsb} lists the conclusions and the argument strengths \asT for each argument for the corresponding four bets involved in the Ellsberg paradox.  

\begin{table}[!ht]
\begin{center} 
\caption{Conclusions and normative strengths (\asTe) of Arguments $\mathcal{A}_1, \ldots, \mathcal{A}_4$  associated with the four bets involved in the Ellsberg paradox. The premises are always $p(R)=.33$ and $p(B \vee Y)=.67$.} 
\label{TAB:ArgEllsb} 
\vskip 0.12in
\begin{tabular}{ccc}\hline
     & Conclusion& Argument strength \\\hline
{\bf Bet 1}& $p(R)=.33$&$\as(\mathcal{A}_1)=.33$ \\
{\bf Bet 2} & $0\leq p(B)\leq.67$&$\as(\mathcal{A}_2)=.11$\\
{\bf Bet 3}& $.33 \leq p(R\vee Y) \leq 1$&$\as(\mathcal{A}_3)=.22$\\
{\bf Bet 4}&{$p(B\vee Y)=.67$}&$\as(\mathcal{A}_4)=.67$\\\hline
\end{tabular}
\end{center}
\end{table}

The four argument strength values in Table~\ref{TAB:ArgEllsb}  induce the following preference orders in the classical Ellsberg task:  {\bf Bet 1} $\succ$  {\bf Bet 2}, since $\as(\mathcal{A}_1)=.33>\as(\mathcal{A}_2)=.11$, and {\bf Bet 4} $\succ$  {\bf Bet 3}, since $\as(\mathcal{A}_4)=.67>\as(\mathcal{A}_3)=.22$ (where $ X \succ Y$ denotes \emph{$X$ is preferred over $Y$}). This preference order corresponds to Ellsberg's predictions and matches the data \cite<see, e.g.,>{becker64,slovic74,maccrimmon79}. 

The functions of the four arguments can be understood in an \emph{epistemic} and in a \emph{persuasive} sense.  The epistemic function of the arguments is to \emph{gain knowledge} about which bet should be preferred. The persuasive function of the arguments is to \emph{convince}  someone which bet should be preferred.

In the following section we further investigate the  psychological plausibility of \asT by an experiment. 

\section{Method}

\subsection{Participants}

In this experiment 60 university students (mean age 25.9 years ($SD=5.6$), 48 females, 12 males)  participated for a compensation of 15\EUR. All of the participants were Finnish native speakers and none of them had studied  psychology, mathematics, statistics or philosophy as their major.

\subsection{Design and Materials}
We used three target task types: argument ranking tasks, argument rating tasks, and the (original) Ellsberg tasks.
The \emph{argument ranking tasks}  first instructed the participants to \emph{rank} the strength of arguments $\mathcal{A}_1$ and $\mathcal{A}_2$ (see  Table~\ref{TAB:ArgEllsb}). Second, the participants were instructed to rank the strength of arguments $\mathcal{A}_3$ and $\mathcal{A}_4$. The \emph{argument rating tasks}  instructed the participants  to \emph{rate} the strength of each of the four arguments. In the original version of the Ellsberg task, participants had to rank which bets they preferred as described in the Introduction. We investigated the following questions which relate argument strength to the Ellsberg problem:
\begin{itemize}
\item Do the results of the argument strength rating tasks predict the responses in the Ellsberg tasks? 
\item  Do the results of the argument strength rating tasks predict the responses in the argument strength ranking tasks? 
\end{itemize} 
Moreover, we explored empirically, whether argument strength formulated in epistemic or in persuasive terms impacts participants' reasoning. Finally, we systematically manipulated the information conveyed in the argument rating and in the argument ranking tasks by the following independent variables: (i) only the uncertainty of the conclusion was presented, (ii) only the uncertainties of the premises were presented, and (iii) uncertainties of the premises and the conclusion were presented. The instructions introduced the following  symbol for marking not conveyed information in the respective conditions which correspond the variables (i) and (ii):  \noiseS. By using a $2\times 3$ between-participant design we fully crossed epistemic versus persuasive formulations and the manipulated information conveyed in the arguments. 
In the epistemic booklets  we used knowledge-oriented phrasings like ``Which argument is stronger to know which bet to choose?'', whereas in the  persuasive booklets we used according phrasings like ``Which argument convinces stronger which bet to choose?''.   The experimental conditions are explained in Table~\ref{TAB:design}. 
\begin{table}[!ht]
\begin{center} 
\caption{Experimental conditions (Cd 1--Cd 6; $N=60$).} 
\label{TAB:design} 
\vskip 0.12in
\begin{tabular}{lll} 
\hline 
Presented probabilities &Epistemic     &  Persuasive  \\\hline
Premise \& conclusion &      Cd 1 ($n_1=10$)        &     Cd 2 ($n_2=10$) \\
Conclusion only&        Cd 3 ($n_3=10$)      &  Cd 4 ($n_4=10$)    \\
Premise only&   Cd 5 ($n_5=10$)           &  Cd 6 ($n_6=10$)    \\\hline
\end{tabular} 
\end{center} 
\end{table}

\paragraph{Argument ranking tasks}
In these tasks, the participants were instructed to imagine two  friends   arguing about which bet the participant  should choose. Then, argument $\mathcal{A}_1$ for {\bf Bet~1}, and argument $\mathcal{A}_2$ for {\bf Bet~2} were presented to the participant, e.g.:
\begin{quote}
\noindent {\bf Argument 2} for {\bf Bet~2}\\
\fbox{
 \begin{minipage}[c][7em][c]{0.43\textwidth}
I am \noiseBOX \% sure that the ball drawn from the urn is red.\\
I am \noiseBOX \% sure that the ball drawn from the urn is black or yellow.\\
{\underline{Therefore}}, I am at least 0 \% and at most 67 \% sure that the ball drawn from the urn is black.
\end{minipage}}
\end{quote}
The participants were then presented with the question ``Which argument is stronger to know which bet to choose?'' (\emph{Kumpi argumentti on vahvempi sen tiet\"amiseen, kumpi veto kannattaisi valita?}) in the epistemic condition. In the persuasive condition, they were asked  \emph{``Which argument convinces you stronger which bet to choose?''} (\emph{Kumpi ar\-gu\-ment\-ti va\-kuut\-taa sinut vah\-vem\-min siit\"a, kumpi veto kan\-nat\-tai\-si valita?}). Then, the participants were instructed to indicate their choice by ticking the respective box for  Argument~1 (i.e., $\mathcal{A}_1$) or  Argument~2 (i.e., $\mathcal{A}_2$). Finally, the participants ranked  Argument~3 (i.e., $\mathcal{A}_3$) and Argument~4 (i.e., $\mathcal{A}_4$).

\begin{figure}[ht]
\begin{center}
\includegraphics[width=.5\textwidth]{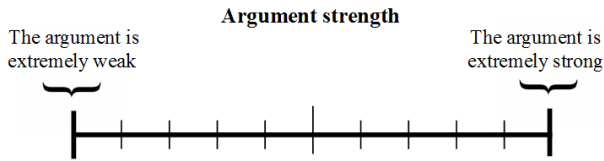} 
\end{center}\vspace{-1cm}
\caption{Answer scale used in the argument rating tasks.} 
\label{argscale}
\end{figure}

\paragraph{Argument rating tasks}
In these tasks participants were presented with the same four arguments as in the argument ranking tasks. They were asked to carefully reconsider each. Instead of using forced choice response formats, each argument was followed by a question, e.g.,   {``How strong is {\bf Argument~2} for choosing {\bf Bet~2}?''} (\emph{Kuinka vahva \textbf{Argumentti 2} on \textbf{Vedon~2} valitsemiseksi?}; original epistemic formulation) or {``How strong is {\bf Argument~2} for convincing to choose {\bf Bet~2}?''} (\emph{Kuinka vahva \textbf{Argumentti 2} on vakuuttamaan \textbf{Vedon~2} va\-lit\-se\-mi\-ses\-ta?}; original persuasive formulation).
The participants were asked to mark their responses on a scale (see Figure~\ref{argscale}).

\paragraph{Ellsberg tasks} Here, as explained in the introduction, the participants had to choose   which rankings among bets they preferred  ({\bf Bet~1} or {\bf Bet~2} and {\bf Bet~3} or {\bf Bet~4}). All   participants were presented with the same Ellsberg tasks.


\subsection{Procedures}

Participants completed the booklets individually in a quiet room. At the beginning of the testing, participants were informed  to take  as much time as  needed for completing the tasks. Furthermore, they were instructed  not to look back on their previous responses. 
After  reading the introduction the participants  worked on  tasks which differed from the Ellsberg problem (and which are not in the scope of the present paper). After that, the target tasks were presented in the following order: (i) argument ranking tasks, (ii) argument rating tasks, and  (iii) the Ellsberg tasks. Finally, the participants filled in demographic data and answered questions about the difficulty and clearness of the tasks. Each session concluded by an interview to further explore argument strength from a qualitative point of view: we asked  how the participants solved the tasks and what they thought determined the strength of an argument. Participants used on the average 9.6 minutes ($SD=2.8$) to work on the target tasks and to fill in the final questions in the booklet. 





\section{Results and Discussion}

We performed Fisher's exact tests to compare the impact of the different booklets on the  response  frequencies in the argument ranking tasks and  in the Ellsberg tasks. Moreover, we tested influences of the different conditions in the argument rating tasks by analyses of variance. After performing Holm-Bonferroni corrections we did not observe any significant differences. We therefore pooled the data for further analysis ($N=60$).

\paragraph{Ellsberg's predictions}
The majority of responses in all three types of tasks  (i.e., argument ranking, argument rating and Ellsberg task) are consistent with Ellsberg's predictions. 
Our findings also replicate  empirical findings reported in the literature \cite<see, e.g.,>{becker64,slovic74,maccrimmon79}. Moreover, our data suggest that classical findings in  Ellsberg tasks  carry over to (isomorphic) problems formulated in terms of argument strength.

Table~\ref{ellsbergfrequence} shows how the participants ranked the arguments in the argument ranking tasks and how they ranked the bets in the Ellsberg tasks. Bet~1 (resp.,  argument $\mathcal{A}_1$ supporting Bet~1) is more frequently chosen than Bet~2 (resp., $\mathcal{A}_2$ supporting Bet~2). Likewise, Bet~4 (resp.,  argument $\mathcal{A}_4$ supporting Bet~4) is more frequently chosen than Bet~3 (resp., $\mathcal{A}_3$ supporting Bet~3).

\begin{table}
\begin{center}
\caption{Percentages of argument preferences in the argument ranking tasks (rnk$(\mathcal{A})$) and in  the Ellsberg tasks ($N=60$).}
\label{ellsbergfrequence}
\vskip 0.12in
\begin{tabular}{ccccccc}
\cline{1-3} \cline{5-7}
\%  &rnk$(\mathcal{A})$   &Ellsberg  & &\%  &rnk$(\mathcal{A})$   &Ellsberg \\ \cline{1-3} \cline{5-7}
Bet1  &73,3    &93,3  &  &Bet3  &25,0    &23,3 \\
Bet2  &26,7    &~6,7 &  &Bet4  &75,0    &76,7 \\ \cline{1-3} \cline{5-7}
\end{tabular}
\end{center}
\end{table}

Moreover, we constructed the underlying preference orders of the  argument strengths and the bets from the participants' responses in all the three task types. This allows one to see which choice strategies were most commonly used. In all  tasks, strategies consistent with the independence axioms of rational choice 
were less frequently preferred, as can be seen in Table~\ref{strategy}. For constructing the preference orders based on the responses in the argument strength ratings tasks, we made the following assumption:   if the strength of an argument $\mathcal{A}_x$ was rated higher than the strength of an argument $\mathcal{A}_y$, then the corresponding Bet~$x$ is preferred over Bet~$y$. Again, our findings replicate the predictions of Ellsberg and the previous empirical findings \cite<see, e.g.,>{becker64,slovic74,maccrimmon79}.


\begin{table}[!ht]
\begin{center} 
\caption{Percentages of responses consistent with  Ellsberg's predictions ($E$), the independence axiom of rational choice ($I$). The preference order $R$ can be interpreted as a reversed version of $E$. ``$(x , y) \succ (u , v)$'' means ``arguments (resp. bets) $x$ and $y$ are preferred over arguments (resp. bets) $u$ and $v$''.  Preference order responses consistent with  \asT are  in \textbf{bold}.}
\label{strategy}
\vskip 0.12in
\begin{tabular}{cccc}
\hline
                       Preference&\multicolumn{3}{c}{Tasks ($N=60$)}\\\cline{2-4}
                       Order&$\mathcal{A}$ Ranking& Ellsberg & $\mathcal{A}$ Rating\\\hline
$(1 , 4) \succ (2 , 3)^E$\!\!\!\!& \textbf{56.67}& \textbf{71.67}& \textbf{56.10}\\
$(2 , 3) \succ (1 , 4)^R$\!\!\!\!&8.33 & 1.67& 4.88 \\
$ (1 , 3) \succ (2 , 4)^I$\!\!\!\!& 16.67& 21.67& 21.95\\
$ (2 , 4) \succ (1 , 3)^I$\!\!\!\!& 18.33& 5.00& 17.07\\\hline
\end{tabular}
\end{center} 
\end{table}


Table~\ref{ellsbergratings} shows the mean argument strength rating responses. As predicted by measure \asTe, the mean argument strength ratings reflect the Ellsberg predictions, i.e., $\text{mean rating}(\mathcal{A}_1)>\text{mean rating}(\mathcal{A}_2)$ and $\text{mean rating}(\mathcal{A}_4)>\text{mean rating}(\mathcal{A}_3)$.

\begin{table}[!ht]
\begin{center} 
\caption{Means and standard deviations ($SD$) of the argument strength  ratings on a scale from 0 (``extremely weak'') to 10 (``extremely strong''; $N=60$).} 
\label{ellsbergratings} 
\vskip 0.12in
\begin{tabular}{ccccc} 
\hline
  &$\mathcal{A}_1$     &  $\mathcal{A}_2$ &$\mathcal{A}_3$   &$\mathcal{A}_4$  \\\hline
Mean &      5,20        &     3,98  &   5,77    &   6,95 \\
$SD$&        2,64      &  2,58    &1,74   &1,87    \\\hline
\end{tabular} 
\end{center} 
\end{table}

\paragraph{Consistency among the data}

Based on the argument strength ratings, we predicted the participants' choices in the ranking and in the Ellsberg tasks. The data support our predictions: the  argument strength rating responses predict the ranking responses in the Ellsberg tasks. The rating responses also predict  the responses in the argument strength ranking tasks (see Table~\ref{prediction1} and Table~\ref{prediction2}).

As some participants had rated the arguments for the bets equally strong, no predictions could be derived in these cases. When taking into account only those cases, in which making predictions was possible, the responses of roughly 3/4 of the participants were consistent with their responses in  the ranking tasks. In the argument strength ranking tasks, 77.3 \% of the participants chose as predicted between the first two bets and 75.0 \%  chose as predicted between the second two bets. For the Ellsberg tasks, we observed similarly high percentages (i.e.,  75.0 \% and 70.8 \% of the participants, for the first and the second bet rankings, respectively). This is again strong experimental support for the psychological plausibility of measure \asTe.



\begin{table}[!ht]
\begin{center} 
\caption{Predictions of bet rankings in Ellsberg tasks based on responses in the argument strength rating tasks ($N=60$).} 
\label{prediction1} 
\vskip 0.12in
\begin{tabular}{lcc} 
\hline
    & \multicolumn{2}{c}{Ranking}\\ \cline{2-3}
 \% & Bet 1 vs. Bet 2\!&Bet 3 vs. Bet 4  \\\hline

Chose as predicted &      55.00        &     56.67 \\
Did not choose as predicted\!\!\!\!\!\!\!\!\!\!\!&        18.33      &  23.33    \\
No prediction made &   26.67          &  20.00    \\\hline
\end{tabular} 
\end{center} 
\end{table}

\begin{table}[!ht]
\begin{center} 
\caption{Predictions of argument strength rankings based on the responses in argument strength rating tasks ($N=60$).} 
\label{prediction2} 
\vskip 0.12in
\begin{tabular}{lcc} 
\hline
    & \multicolumn{2}{c}{Ranking}\\ \cline{2-3}
 \% & $\mathcal{A}_1$ vs. $\mathcal{A}_2$     &$\mathcal{A}_3$  vs. $\mathcal{A}_4$  \\\hline
Chose as predicted &      56.67        &     60.00 \\
Did not choose as predicted&        16.67      &  20.00    \\
No prediction made &   26.67          &  20.00    \\\hline
\end{tabular} 
\end{center} 
\end{table}

Finally, we discuss qualitative data taken from structured interviews on folk psychological conceptions on what argument strength means.

\paragraph{Interview results} After the participants completed the paper and pencil tasks, we collected folk psychological conceptions on what ``argument strength'' (\emph{argumentin vahvuus}) means by structured interviews.
We asked the participants how they would define argument strength in their own words. Participants who had received the \emph{persuasive} booklets, we hypothesized,  mentioned persuasive aspects (like how \emph{convincing}  arguments are)  more frequently than those of the epistemic condition. Moreover, participants who had received the  \emph{epistemic} booklets  focused more on epistemic aspects (like truth and knowledge) than those of the persuasive condition. However, the interview responses do not   confirm these hypotheses.

The responses to the interview question concerning  the meaning of ``argument strength'' reflected features of our measure  \asTe.
Specifically, the \emph{location} of the coherent conclusion probability interval was referred to by almost all of the participants. 
For many participants the location seemed to be more important than the \emph{precision} of the coherent conclusion probability interval. They had, for example, focused solely on the lower probability bound of the interval and ignored the upper bound or responded based on the mean value of the interval. 
 

However, a few participants also referred to the \emph{precision} of the coherent conclusion probability interval by sentences like:
\begin{quote}
``The size of this gap between 33 [\%] and 100 [\%] is so big that it increases the uncertainty.'' \emph{(Ep\"avarmuutta lis\"a\"a se, ett\"a v\"ali 33:n ja 100:n v\"alill\"a on niin suuri)}
\end{quote}
Some participants also talked about the truth or correctness of the probability bounds of the conclusion.  For them, the arguments were strong, when the probabilities in the conclusions were \emph{correct}, almost regardless of the values in them. 

Finally, we note that the interview responses provide folk psychological evidence for using location and precision of  conclusion probability intervals for evaluating the strength of uncertain arguments. Location and precision  are  the key ingredients of our  measure of argument strength~\asTe.






\section{Concluding Remarks}

We proposed a formal measure of argument strength and showed how it predicts  responses in Ellsberg tasks. Specifically, we framed choices among bets in terms of probability logical argument forms. We confirmed experimentally that  Ellsberg's predictions can be justified by argument strength rankings and argument strength ratings.

Since the proposed measure exploits tools available in coherence-based probability logic and since it is based on  a deductive consequence relation, it allows for dealing with arguments involving conditionals. The proposed measure has many plausible consequences, which calls for future formal-normative and experimental research for modeling also other argument types, like the conditional syllogisms.

Understanding argument strength is important for theories about reasoning and argumentation in general. Our paper sheds formal and experimental light on what  argument strength means.

\section{Acknowledgments}

This research was supported by the DFG project
    PF~740/2-2 (awarded to Niki Pfeifer) as part of the Priority Program ``New Frameworks of Rationality'' (SPP1516).




\begin{thebibliography}{}

\bibitem [\protect \citeauthoryear {%
Becker%
\ \BBA {} Brownson%
}{%
Becker%
\ \BBA {} Brownson%
}{%
{\protect \APACyear {1964}}%
}]{%
becker64}
\APACinsertmetastar {%
becker64}%
\begin{APACrefauthors}%
Becker, S\BPBI W.%
\BCBT {}\ \BBA {} Brownson, F\BPBI O.%
\end{APACrefauthors}%
\unskip\
\newblock
\APACrefYearMonthDay{1964}{}{}.
\newblock
{\BBOQ}\APACrefatitle {What Price Ambiguity? or the Role of Ambiguity in
  Decision-Making} {What price ambiguity? or the role of ambiguity in
  decision-making}.{\BBCQ}
\newblock
\APACjournalVolNumPages{Journal of Political Economy}{72}{1}{62--73}.
\PrintBackRefs{\CurrentBib}

\bibitem [\protect \citeauthoryear {%
Briggs%
}{%
Briggs%
}{%
{\protect \APACyear {2016}}%
}]{%
sep-rationality-normative-utility}
\APACinsertmetastar {%
sep-rationality-normative-utility}%
\begin{APACrefauthors}%
Briggs, R.%
\end{APACrefauthors}%
\unskip\
\newblock
\APACrefYearMonthDay{2016}{}{}.
\newblock
{\BBOQ}\APACrefatitle {Normative Theories of Rational Choice: {E}xpected
  Utility} {Normative theories of rational choice: {E}xpected utility}.{\BBCQ}
\newblock
\BIn{} E\BPBI N.~Zalta\ (\BED), \APACrefbtitle {The {S}tanford Encyclopedia of
  Philosophy} {The {S}tanford encyclopedia of philosophy}\
  (\PrintOrdinal{Winter 2016}\ \BEd).
\newblock
\APAChowpublished {\url{http://tinyurl.com/hjwmajw}}.
\PrintBackRefs{\CurrentBib}

\bibitem [\protect \citeauthoryear {%
Coletti%
\ \BBA {} Scozzafava%
}{%
Coletti%
\ \BBA {} Scozzafava%
}{%
{\protect \APACyear {2002}}%
}]{%
coletti02}
\APACinsertmetastar {%
coletti02}%
\begin{APACrefauthors}%
Coletti, G.%
\BCBT {}\ \BBA {} Scozzafava, R.%
\end{APACrefauthors}%
\unskip\
\newblock
\APACrefYear{2002}.
\newblock
\APACrefbtitle {Probabilistic logic in a coherent setting} {Probabilistic logic
  in a coherent setting}.
\newblock
\APACaddressPublisher{Dordrecht}{Kluwer}.
\PrintBackRefs{\CurrentBib}

\bibitem [\protect \citeauthoryear {%
Crupi%
, Tentori%
\BCBL {}\ \BBA {} Gonzales%
}{%
Crupi%
\ \protect \BOthers {.}}{%
{\protect \APACyear {2007}}%
}]{%
crupi07}
\APACinsertmetastar {%
crupi07}%
\begin{APACrefauthors}%
Crupi, V.%
, Tentori, K.%
\BCBL {}\ \BBA {} Gonzales, M.%
\end{APACrefauthors}%
\unskip\
\newblock
\APACrefYearMonthDay{2007}{}{}.
\newblock
{\BBOQ}\APACrefatitle {On {B}ayesian measures of confirmation} {On {B}ayesian
  measures of confirmation}.{\BBCQ}
\newblock
\APACjournalVolNumPages{Philosophy of Science}{74}{}{229-252}.
\PrintBackRefs{\CurrentBib}

\bibitem [\protect \citeauthoryear {%
{de~Finetti}%
}{%
{de~Finetti}%
}{%
{\protect \APACyear {1970/1974}}%
}]{%
definetti74}
\APACinsertmetastar {%
definetti74}%
\begin{APACrefauthors}%
{de~Finetti}, B.%
\end{APACrefauthors}%
\unskip\
\newblock
\APACrefYear{1970/1974}.
\newblock
\APACrefbtitle {Theory of probability} {Theory of probability}\ (\BVOLS\ 1, 2).
\newblock
\APACaddressPublisher{Chichester}{John Wiley \& Sons}.
\PrintBackRefs{\CurrentBib}

\bibitem [\protect \citeauthoryear {%
Ellsberg%
}{%
Ellsberg%
}{%
{\protect \APACyear {1961}}%
}]{%
ellsberg61}
\APACinsertmetastar {%
ellsberg61}%
\begin{APACrefauthors}%
Ellsberg, D.%
\end{APACrefauthors}%
\unskip\
\newblock
\APACrefYearMonthDay{1961}{}{}.
\newblock
{\BBOQ}\APACrefatitle {Risk, ambiguity, and the {S}avage axioms} {Risk,
  ambiguity, and the {S}avage axioms}.{\BBCQ}
\newblock
\APACjournalVolNumPages{The Quarterly Journal of Economics}{75}{4}{643--669}.
\PrintBackRefs{\CurrentBib}

\bibitem [\protect \citeauthoryear {%
Evans%
\ \BBA {} Over%
}{%
Evans%
\ \BBA {} Over%
}{%
{\protect \APACyear {2004}}%
}]{%
evans04}
\APACinsertmetastar {%
evans04}%
\begin{APACrefauthors}%
Evans, J\BPBI {\relax St}\BPBI B\BPBI T.%
\BCBT {}\ \BBA {} Over, D\BPBI E.%
\end{APACrefauthors}%
\unskip\
\newblock
\APACrefYear{2004}.
\newblock
\APACrefbtitle {If} {If}.
\newblock
\APACaddressPublisher{Oxford}{Oxford University Press}.
\PrintBackRefs{\CurrentBib}

\bibitem [\protect \citeauthoryear {%
Gilio%
, Over%
, Pfeifer%
\BCBL {}\ \BBA {} Sanfilippo%
}{%
Gilio%
\ \protect \BOthers {.}}{%
{\protect \APACyear {2017}}%
}]{%
2016:SMPS1}
\APACinsertmetastar {%
2016:SMPS1}%
\begin{APACrefauthors}%
Gilio, A.%
, Over, D\BPBI E.%
, Pfeifer, N.%
\BCBL {}\ \BBA {} Sanfilippo, G.%
\end{APACrefauthors}%
\unskip\
\newblock
\APACrefYearMonthDay{2017}{}{}.
\newblock
{\BBOQ}\APACrefatitle {Centering and compound conditionals under coherence}
  {Centering and compound conditionals under coherence}.{\BBCQ}
\newblock
\BIn{} M\BPBI B.~Ferraro\ \BOthers {.}\ (\BEDS), \APACrefbtitle {Soft Methods
  for Data Science} {Soft methods for data science}\ (\BPGS\ 253--260).
\newblock
\APACaddressPublisher{Berlin, Heidelberg}{Springer}.
\PrintBackRefs{\CurrentBib}

\bibitem [\protect \citeauthoryear {%
Gilio%
, Over%
, Pfeifer%
\BCBL {}\ \BBA {} Sanfilippo%
}{%
Gilio%
\ \protect \BOthers {.}}{%
{\protect \APACyear {submitted}}%
}]{%
GOPSsubm}
\APACinsertmetastar {%
GOPSsubm}%
\begin{APACrefauthors}%
Gilio, A.%
, Over, D\BPBI E.%
, Pfeifer, N.%
\BCBL {}\ \BBA {} Sanfilippo, G.%
\end{APACrefauthors}%
\unskip\
\newblock
\APACrefYearMonthDay{submitted}{}{}.
\newblock
\APACrefbtitle {Centering with conjoined and iterated conditionals under
  coherence.} {Centering with conjoined and iterated conditionals under
  coherence.}
\newblock
\APAChowpublished {https://arxiv.org/abs/1701.07785}.
\PrintBackRefs{\CurrentBib}

\bibitem [\protect \citeauthoryear {%
Gilio%
, Pfeifer%
\BCBL {}\ \BBA {} Sanfilippo%
}{%
Gilio%
\ \protect \BOthers {.}}{%
{\protect \APACyear {2016}}%
}]{%
gilio16}
\APACinsertmetastar {%
gilio16}%
\begin{APACrefauthors}%
Gilio, A.%
, Pfeifer, N.%
\BCBL {}\ \BBA {} Sanfilippo, G.%
\end{APACrefauthors}%
\unskip\
\newblock
\APACrefYearMonthDay{2016}{}{}.
\newblock
{\BBOQ}\APACrefatitle {Transitivity in coherence-based probability logic.}
  {Transitivity in coherence-based probability logic.}{\BBCQ}
\newblock
\APACjournalVolNumPages{Journal of Applied Logic}{14}{}{46--64}.
\PrintBackRefs{\CurrentBib}

\bibitem [\protect \citeauthoryear {%
Haenni%
}{%
Haenni%
}{%
{\protect \APACyear {2009}}%
}]{%
haenni09}
\APACinsertmetastar {%
haenni09}%
\begin{APACrefauthors}%
Haenni, R.%
\end{APACrefauthors}%
\unskip\
\newblock
\APACrefYearMonthDay{2009}{}{}.
\newblock
{\BBOQ}\APACrefatitle {Probabilistic argumentation} {Probabilistic
  argumentation}.{\BBCQ}
\newblock
\APACjournalVolNumPages{Journal of Applied Logic}{}{}{155--176}.
\PrintBackRefs{\CurrentBib}

\bibitem [\protect \citeauthoryear {%
Hahn%
\ \BBA {} Oaksford%
}{%
Hahn%
\ \BBA {} Oaksford%
}{%
{\protect \APACyear {2006}}%
}]{%
hahn06a}
\APACinsertmetastar {%
hahn06a}%
\begin{APACrefauthors}%
Hahn, U.%
\BCBT {}\ \BBA {} Oaksford, M.%
\end{APACrefauthors}%
\unskip\
\newblock
\APACrefYearMonthDay{2006}{}{}.
\newblock
{\BBOQ}\APACrefatitle {A normative theory of argument strength} {A normative
  theory of argument strength}.{\BBCQ}
\newblock
\APACjournalVolNumPages{Informal Logic}{26}{}{1-22}.
\PrintBackRefs{\CurrentBib}

\bibitem [\protect \citeauthoryear {%
Mac{C}rimmon%
\ \BBA {} Larsson%
}{%
Mac{C}rimmon%
\ \BBA {} Larsson%
}{%
{\protect \APACyear {1979}}%
}]{%
maccrimmon79}
\APACinsertmetastar {%
maccrimmon79}%
\begin{APACrefauthors}%
Mac{C}rimmon, K\BPBI R.%
\BCBT {}\ \BBA {} Larsson, S.%
\end{APACrefauthors}%
\unskip\
\newblock
\APACrefYearMonthDay{1979}{}{}.
\newblock
{\BBOQ}\APACrefatitle {Utility theory: {A}xioms versus `paradoxes'} {Utility
  theory: {A}xioms versus `paradoxes'}.{\BBCQ}
\newblock
\BIn{} M.~Allais\ \BBA {} O.~Hagen\ (\BEDS), \APACrefbtitle {Expected Utility
  and the {A}llais Paradox} {Expected utility and the {A}llais paradox}\
  (\BVOL\ 1979, \BPGS\ 333--409).
\newblock
\APACaddressPublisher{Dordrecht}{Reidel}.
\PrintBackRefs{\CurrentBib}

\bibitem [\protect \citeauthoryear {%
Oaksford%
\ \BBA {} Chater%
}{%
Oaksford%
\ \BBA {} Chater%
}{%
{\protect \APACyear {2007}}%
}]{%
oaksford07}
\APACinsertmetastar {%
oaksford07}%
\begin{APACrefauthors}%
Oaksford, M.%
\BCBT {}\ \BBA {} Chater, N.%
\end{APACrefauthors}%
\unskip\
\newblock
\APACrefYear{2007}.
\newblock
\APACrefbtitle {Bayesian rationality: {T}he probabilistic approach to human
  reasoning} {Bayesian rationality: {T}he probabilistic approach to human
  reasoning}.
\newblock
\APACaddressPublisher{Oxford}{Oxford University Press}.
\PrintBackRefs{\CurrentBib}

\bibitem [\protect \citeauthoryear {%
Oaksford%
\ \BBA {} Hahn%
}{%
Oaksford%
\ \BBA {} Hahn%
}{%
{\protect \APACyear {2007}}%
}]{%
oaksford07c}
\APACinsertmetastar {%
oaksford07c}%
\begin{APACrefauthors}%
Oaksford, M.%
\BCBT {}\ \BBA {} Hahn, U.%
\end{APACrefauthors}%
\unskip\
\newblock
\APACrefYearMonthDay{2007}{}{}.
\newblock
{\BBOQ}\APACrefatitle {Induction, deduction, and argument strength in human
  reasoning and argumentation} {Induction, deduction, and argument strength in
  human reasoning and argumentation}.{\BBCQ}
\newblock
\BIn{} A.~Feeney\ \BBA {} E.~Heit\ (\BEDS), \APACrefbtitle {Inductive
  reasoning. {E}xperimental, developmental, and computational approaches}
  {Inductive reasoning. {E}xperimental, developmental, and computational
  approaches}\ (\BPGS\ 269--301).
\newblock
\APACaddressPublisher{Cambridge}{Cambridge University Press}.
\PrintBackRefs{\CurrentBib}

\bibitem [\protect \citeauthoryear {%
Osherson%
, Smith%
, Wilkie%
, L\'opez%
\BCBL {}\ \BBA {} Shafir%
}{%
Osherson%
\ \protect \BOthers {.}}{%
{\protect \APACyear {1990}}%
}]{%
osherson90}
\APACinsertmetastar {%
osherson90}%
\begin{APACrefauthors}%
Osherson, D\BPBI N.%
, Smith, E\BPBI E.%
, Wilkie, O.%
, L\'opez, A.%
\BCBL {}\ \BBA {} Shafir, E.%
\end{APACrefauthors}%
\unskip\
\newblock
\APACrefYearMonthDay{1990}{}{}.
\newblock
{\BBOQ}\APACrefatitle {Category-based induction} {Category-based
  induction}.{\BBCQ}
\newblock
\APACjournalVolNumPages{Psychological Review}{97}{2}{185--200}.
\PrintBackRefs{\CurrentBib}

\bibitem [\protect \citeauthoryear {%
Over%
\ \BBA {} Cruz%
}{%
Over%
\ \BBA {} Cruz%
}{%
{\protect \APACyear {{\protect \BIP {}}}}%
}]{%
OverCruz17}
\APACinsertmetastar {%
OverCruz17}%
\begin{APACrefauthors}%
Over, D\BPBI E.%
\BCBT {}\ \BBA {} Cruz, N.%
\end{APACrefauthors}%
\unskip\
\newblock
\APACrefYearMonthDay{{\protect \BIP {}}}{}{}.
\newblock
{\BBOQ}\APACrefatitle {Probabilistic accounts of conditional reasoning}
  {Probabilistic accounts of conditional reasoning}.{\BBCQ}
\newblock
\BIn{} L.~Macchi, M.~Bagassi\BCBL {}\ \BBA {} R.~Vialem\ (\BEDS),
  \APACrefbtitle {International Handbook of Thinking and Reasoning.}
  {International handbook of thinking and reasoning.}
\newblock
\APACaddressPublisher{Hove Sussex}{Psychology Press}.
\PrintBackRefs{\CurrentBib}

\bibitem [\protect \citeauthoryear {%
Pfeifer%
}{%
Pfeifer%
}{%
{\protect \APACyear {2007}}%
}]{%
pfeifer07}
\APACinsertmetastar {%
pfeifer07}%
\begin{APACrefauthors}%
Pfeifer, N.%
\end{APACrefauthors}%
\unskip\
\newblock
\APACrefYearMonthDay{2007}{}{}.
\newblock
{\BBOQ}\APACrefatitle {Rational argumentation under uncertainty} {Rational
  argumentation under uncertainty}.{\BBCQ}
\newblock
\BIn{} G.~Kreuzbauer, N.~Gratzl\BCBL {}\ \BBA {} E.~Hiebl\ (\BEDS),
  \APACrefbtitle {Persuasion und {W}issenschaft: {A}ktuelle {F}ragestellungen
  von {R}hetorik und {A}rgumentationstheorie} {Persuasion und {W}issenschaft:
  {A}ktuelle {F}ragestellungen von {R}hetorik und {A}rgumentationstheorie}\
  (\BPGS\ 181--191).
\newblock
\APACaddressPublisher{Wien}{{\sc Lit} Verlag}.
\PrintBackRefs{\CurrentBib}

\bibitem [\protect \citeauthoryear {%
Pfeifer%
}{%
Pfeifer%
}{%
{\protect \APACyear {2013}}%
{\protect \APACexlab {{\protect \BCnt {1}}}}}]{%
pfeifer13b}
\APACinsertmetastar {%
pfeifer13b}%
\begin{APACrefauthors}%
Pfeifer, N.%
\end{APACrefauthors}%
\unskip\
\newblock
\APACrefYearMonthDay{2013{\protect \BCnt {1}}}{}{}.
\newblock
{\BBOQ}\APACrefatitle {The new psychology of reasoning: {A} mental probability
  logical perspective} {The new psychology of reasoning: {A} mental probability
  logical perspective}.{\BBCQ}
\newblock
\APACjournalVolNumPages{Thinking \& Reasoning}{19}{3--4}{329--345}.
\PrintBackRefs{\CurrentBib}

\bibitem [\protect \citeauthoryear {%
Pfeifer%
}{%
Pfeifer%
}{%
{\protect \APACyear {2013}}%
{\protect \APACexlab {{\protect \BCnt {2}}}}}]{%
pfeifer13a}
\APACinsertmetastar {%
pfeifer13a}%
\begin{APACrefauthors}%
Pfeifer, N.%
\end{APACrefauthors}%
\unskip\
\newblock
\APACrefYearMonthDay{2013{\protect \BCnt {2}}}{}{}.
\newblock
{\BBOQ}\APACrefatitle {On argument strength} {On argument strength}.{\BBCQ}
\newblock
\BIn{} F.~Zenker\ (\BED), \APACrefbtitle {Bayesian argumentation. {T}he
  practical side of probability} {Bayesian argumentation. {T}he practical side
  of probability}\ (\BPGS\ 185--193).
\newblock
\APACaddressPublisher{Dordrecht}{Synthese Library (Springer)}.
\PrintBackRefs{\CurrentBib}

\bibitem [\protect \citeauthoryear {%
Pfeifer%
}{%
Pfeifer%
}{%
{\protect \APACyear {2014}}%
}]{%
pfeifer13}
\APACinsertmetastar {%
pfeifer13}%
\begin{APACrefauthors}%
Pfeifer, N.%
\end{APACrefauthors}%
\unskip\
\newblock
\APACrefYearMonthDay{2014}{}{}.
\newblock
{\BBOQ}\APACrefatitle {Reasoning about uncertain conditionals} {Reasoning about
  uncertain conditionals}.{\BBCQ}
\newblock
\APACjournalVolNumPages{Studia Logica}{102}{4}{849-866}.
\PrintBackRefs{\CurrentBib}

\bibitem [\protect \citeauthoryear {%
Pfeifer%
\ \BBA {} Kleiter%
}{%
Pfeifer%
\ \BBA {} Kleiter%
}{%
{\protect \APACyear {2006}}%
}]{%
pfeifer06d}
\APACinsertmetastar {%
pfeifer06d}%
\begin{APACrefauthors}%
Pfeifer, N.%
\BCBT {}\ \BBA {} Kleiter, G\BPBI D.%
\end{APACrefauthors}%
\unskip\
\newblock
\APACrefYearMonthDay{2006}{}{}.
\newblock
{\BBOQ}\APACrefatitle {Inference in conditional probability logic} {Inference
  in conditional probability logic}.{\BBCQ}
\newblock
\APACjournalVolNumPages{Kybernetika}{42}{}{391-404}.
\PrintBackRefs{\CurrentBib}

\bibitem [\protect \citeauthoryear {%
Pfeifer%
\ \BBA {} Kleiter%
}{%
Pfeifer%
\ \BBA {} Kleiter%
}{%
{\protect \APACyear {2009}}%
}]{%
pfeifer09b}
\APACinsertmetastar {%
pfeifer09b}%
\begin{APACrefauthors}%
Pfeifer, N.%
\BCBT {}\ \BBA {} Kleiter, G\BPBI D.%
\end{APACrefauthors}%
\unskip\
\newblock
\APACrefYearMonthDay{2009}{}{}.
\newblock
{\BBOQ}\APACrefatitle {Framing human inference by coherence based probability
  logic} {Framing human inference by coherence based probability logic}.{\BBCQ}
\newblock
\APACjournalVolNumPages{Journal of Applied Logic}{7}{2}{206--217}.
\PrintBackRefs{\CurrentBib}

\bibitem [\protect \citeauthoryear {%
Prakken%
\ \BBA {} Vreeswijk%
}{%
Prakken%
\ \BBA {} Vreeswijk%
}{%
{\protect \APACyear {2002}}%
}]{%
prakken02}
\APACinsertmetastar {%
prakken02}%
\begin{APACrefauthors}%
Prakken, H.%
\BCBT {}\ \BBA {} Vreeswijk, G.%
\end{APACrefauthors}%
\unskip\
\newblock
\APACrefYearMonthDay{2002}{}{}.
\newblock
{\BBOQ}\APACrefatitle {Logic for defeasible argumentation} {Logic for
  defeasible argumentation}.{\BBCQ}
\newblock
\BIn{} D\BPBI M.~Gabbay\ \BBA {} F.~Guenthner\ (\BEDS), \APACrefbtitle
  {Handbook of Philosophical Logic} {Handbook of philosophical logic}\
  (\PrintOrdinal{2\textsuperscript{nd}}\ \BEd, \BVOL~4, \BPGS\ 219--318).
\newblock
\APACaddressPublisher{Dordrecht}{Kluwer}.
\PrintBackRefs{\CurrentBib}

\bibitem [\protect \citeauthoryear {%
Slovic%
\ \BBA {} Tversky%
}{%
Slovic%
\ \BBA {} Tversky%
}{%
{\protect \APACyear {1974}}%
}]{%
slovic74}
\APACinsertmetastar {%
slovic74}%
\begin{APACrefauthors}%
Slovic, P.%
\BCBT {}\ \BBA {} Tversky, A.%
\end{APACrefauthors}%
\unskip\
\newblock
\APACrefYearMonthDay{1974}{}{}.
\newblock
{\BBOQ}\APACrefatitle {Who Accepts {S}avage's Axiom?} {Who accepts {S}avage's
  axiom?}{\BBCQ}
\newblock
\APACjournalVolNumPages{Behavioral Science}{19}{6}{368--373}.
\PrintBackRefs{\CurrentBib}

\bibitem [\protect \citeauthoryear {%
Zenker%
}{%
Zenker%
}{%
{\protect \APACyear {2013}}%
}]{%
zenker13}
\APACinsertmetastar {%
zenker13}%
\begin{APACrefauthors}%
Zenker, F.%
\end{APACrefauthors}%
\ (\BED).
\unskip\
\newblock
\APACrefYear{2013}.
\newblock
\APACrefbtitle {Bayesian argumentation: {T}he practical side of probability}
  {Bayesian argumentation: {T}he practical side of probability}.
\newblock
\APACaddressPublisher{Dordrecht}{Synthese Library (Springer)}.
\PrintBackRefs{\CurrentBib}

\end{thebibliography}

\end{document}